\documentclass[10pt,twocolumn,letterpaper]{article}

\usepackage{cvpr}
\usepackage{times}
\usepackage{epsfig}
\usepackage{graphicx}
\usepackage{amsmath}
\usepackage{amssymb}
\usepackage[dvipsnames]{xcolor}
\usepackage{fleqn, tabularx, wrapfig}
\usepackage{enumerate}
\usepackage{subcaption}
\usepackage{booktabs}
\usepackage{multirow}
\usepackage[export]{adjustbox}
\usepackage{enumitem}

\usepackage[breaklinks=true, bookmarks=false]{hyperref}
\definecolor{darkblue}{rgb}{0.2, 0.3, 0.53}
\definecolor{wine}{rgb}{0.5, 0.04, 0.1}
\hypersetup{
     colorlinks = true,
     citecolor = darkblue,
     linkcolor = wine
}

\makeatletter
\newcommand{\mybox}{%
    \collectbox{%
        \setlength{\fboxsep}{1pt}%
        \fbox{\BOXCONTENT}%
    }%
}
\makeatother

\newcolumntype{Y}{>{\centering\arraybackslash}X}
\makeatletter
\newcommand\notsotiny{\@setfontsize\notsotiny{7}{8}}
\makeatother

\cvprfinalcopy 

\def\cvprPaperID{829} 

\ifcvprfinal\fi
\begin{document}

\title{End-to-End Multi-Task Learning with Attention}

\author{Shikun Liu\qquad Edward Johns \qquad Andrew J. Davison\\
Department of Computing, Imperial College London\\
{\tt\small \{shikun.liu17, e.johns, a.davison\}@imperial.ac.uk}
}

\maketitle

\begin{abstract}
   We propose a novel multi-task learning architecture, which allows learning of task-specific feature-level attention. Our design, the Multi-Task Attention Network (MTAN), consists of a single shared network containing a global feature pool, together with a soft-attention module for each task. These modules allow for learning of task-specific features from the global features, whilst simultaneously allowing for features to be shared across different tasks. The architecture can be trained end-to-end and can be built upon any feed-forward neural network, is simple to implement, and is parameter efficient. We evaluate our approach on a variety of datasets, across both image-to-image predictions and image classification tasks. We show that our architecture is state-of-the-art in multi-task learning compared to existing methods, and is also less sensitive to various weighting schemes in the multi-task loss function. Code is available at \url{https://github.com/lorenmt/mtan}.
\end{abstract}

\section{Introduction}
Convolutional Neural Networks (CNNs) have seen great success in a range of computer vision tasks, including image classification \cite{he2016deep}, semantic segmentation \cite{badrinarayanan2015segnet}, and style transfer \cite{johnson2016perceptual}. However, these networks are typically designed to achieve only one particular task. For more complete vision systems in real-world applications, a network which can perform multiple tasks simultaneously is far more desirable than building a set of independent networks, one for each task. This is more efficient not only in terms of memory and inference speed, but also in terms of data, since related tasks may share informative visual features.

This type of learning is called Multi-Task Learning (MTL) \cite{misra2016cross,kendall2017multi,Doersch_2017_ICCV}, and in this paper we present a novel architecture for MTL based on feature-level attention masks, which add greater flexibility to share complementary features. Compared to standard single-task learning, training multiple tasks whilst successfully learning a shared representation poses two key challenges:

\begin{enumerate}[leftmargin=1.2em,label=\roman*)]
  \setlength\itemsep{0em}
  \item {\bf Network Architecture (how to share)}: A multi-task learning architecture should express both {\it task-shared} and {\it task-specific} features. In this way, the network is encouraged to learn a generalisable representation (to avoid over-fitting), whilst also providing the ability to learn features tailored to each task (to avoid under-fitting).
  \item {\bf Loss Function (how to balance tasks)}: A multi-task loss function, which weights the relative contributions of each task, should enable learning of all tasks with equal importance, without allowing easier tasks to dominate. Manual tuning of loss weights is tedious, and it is preferable to automatically learn the weights, or design a network which is robust to different weights.
\end{enumerate}

\begin{figure}[t!]
    \centering
    \includegraphics[width=\linewidth]{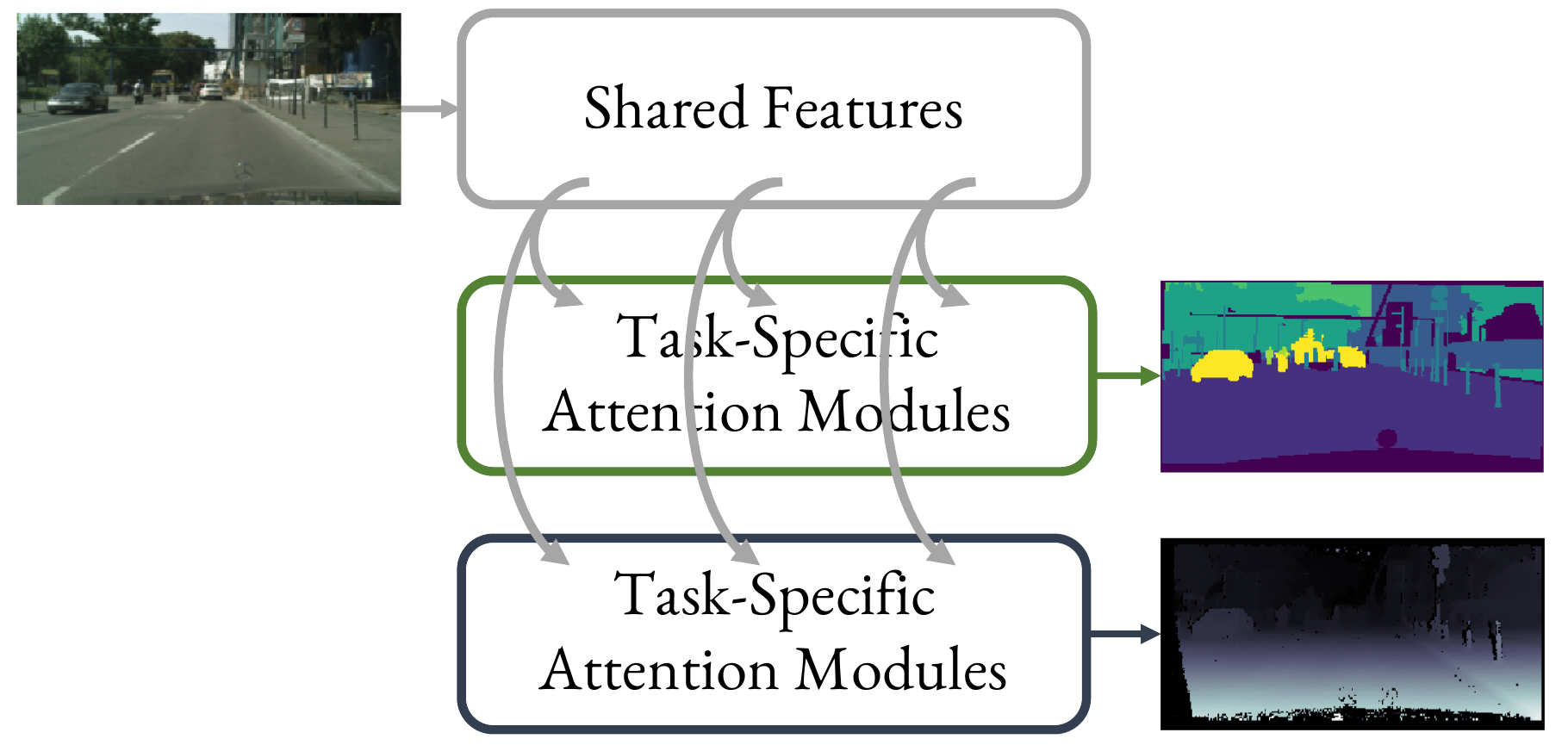}
  \caption{Overview of our proposal MTAN. The shared network takes input data and learns task-shared features, whilst each attention network learns task-specific features, by applying attention modules to the shared network.}
  \label{fig:intro}
  \vspace{-0.25cm}
\end{figure}

However, most prior MTL approaches focus on only one of these two challenges, whilst maintaining a standard implementation of the other. In this paper, we introduce a unified approach which addresses both challenges cohesively, by designing a novel network which (i) enables both task-shared and task-specific features to be learned automatically, and consequently (ii) learns an inherent robustness to the choice of loss weighting scheme.

The proposed network, which we call the Multi-Task Attention Network (MTAN) (see Figure \ref{fig:intro}), is composed of a single shared network, which learns a global feature pool containing features across all tasks. Then for each task, rather than learning directly from the shared feature pool, a soft attention mask is applied at each convolution block in the shared network. In this way, each attention mask automatically determines the importance of the shared features for the respective task, allowing learning of both task-shared and task-specific features in a self-supervised, end-to-end manner. This flexibility enables much more expressive combinations of features to be learned for generalisation across tasks, whilst still allowing for discriminative features to be tailored for each individual task. Furthermore, automatically choosing which features to share and which to be task specific allows for a highly efficient architecture with far fewer parameters than multi-task architectures which have explicit separation of tasks \cite{rusu2016progressive,misra2016cross}.

MTAN can be built on any feed-forward neural network depending on the type of tasks. We first evaluate MTAN with SegNet \cite{badrinarayanan2015segnet}, an encoder-decoder network on the tasks of semantic segmentation and depth estimation on the outdoor CityScapes dataset \cite{cordts2016CityScapes}, and then with an additional task of surface normal prediction on the more challenging indoor dataset NYUv2 \cite{Silberman:ECCV12}. We also test our approach with a different backbone architecture, Wide Residual Network \cite{BMVCwrn}, on the recently proposed Visual Decathlon Challenge \cite{rebuffi2017learning}, to solve 10 individual image classification tasks. Results show that MTAN outperforms several baselines and is competitive with the state-of-the-art for multi-task learning, whilst being more parameter efficient and therefore scaling more gracefully with the number of tasks. Furthermore, our method shows greater robustness to the choice of weighting scheme in the loss function compared to baselines. As part of our evaluation of this robustness, we also propose a novel weighting scheme, Dynamic Weight Average (DWA), which adapts the task weighting over time by considering the rate of change of the loss for each task.



\section{Related Work}

The term Multi-Task Learning (MTL) has been broadly used in machine learning \cite{caruana1998multitask,evgeniou2004regularized,Doersch_2017_ICCV,kumar2012learning}, with similarities to transfer learning \cite{pan2010survey,long2013transfer} and continual learning \cite{thrun2012learning}. In computer vision, multi-task learning has been used to for learning similar tasks such as image classification in multiple domains \cite{rebuffi2017learning}, pose estimation and action recognition \cite{gkioxari2014r}, and dense prediction of depth, surface normals, and semantic classes \cite{misra2016cross,eigen2015predicting}. In this paper, we consider two important aspects of multi-task learning: how can a good multi-task network architecture be designed, and how to balance feature sharing in multi-task learning across all tasks?

Most multi-task learning network architectures for computer vision are designed based on existing CNN architectures. For example, Cross-Stitch Networks \cite{misra2016cross} contain one standard feed-forward network per task, with cross-stitch units to allow features to be shared across tasks. The self-supervised approach of \cite{Doersch_2017_ICCV}, based on the ResNet101 architecture \cite{wang2017residual}, learns a regularised combination of features from different layers of a single shared network. UberNet \cite{Kokkinos_2017_CVPR} proposes an image pyramid approach to process images across multiple resolutions, where for each resolution, additional task-specific layers are formed top of the shared VGG-Net \cite{simonyan2014very}. The Progressive Networks \cite{rusu2016progressive} uses a sequence of incrementally-trained networks to transfer knowledge between tasks. However, architectures such as Cross-Stitch Networks and Progressive Networks require a large number of network parameters, and scale linearly with the number of tasks. In contrast, our model requires only a rough 10\% increase in parameters for per learning task.

On the balancing of feature sharing in multi-task learning, there is extensive experimental analysis in \cite{misra2016cross,kendall2017multi}, with both papers arguing that different amounts of sharing and weighting tend to work best for different tasks. One example of weighting tasks appropriately is with the use of weight uncertainty \cite{kendall2017multi}, which modifies the loss functions in multi-task learning using task uncertainty. Another method is that of GradNorm \cite{chen2017gradnorm}, which manipulates gradient norms over time to control the training dynamics. As an alternative to using task losses to determine task difficulties, Dynamic Task Prioritisation \cite{guo2018dynamic} encourages prioritisation of difficult tasks directly using performance metrics such as accuracy and precision.

\section{Multi-Task Attention Network}
We now introduce our novel multi-task learning architecture, the Multi-Task Attention Network (MTAN). Whilst the architecture can be incorporated into any feed-forward network, in the following section we demonstrate how to build MTAN upon an encoder-decoder network, SegNet \cite{badrinarayanan2015segnet}. This example configuration allows for image-to-image dense pixel-level prediction, such as semantic segmentation, depth estimation, and surface normal prediction.

\subsection{Architecture Design}
\begin{figure*}[ht!]
  \centering
  \includegraphics[width=\textwidth]{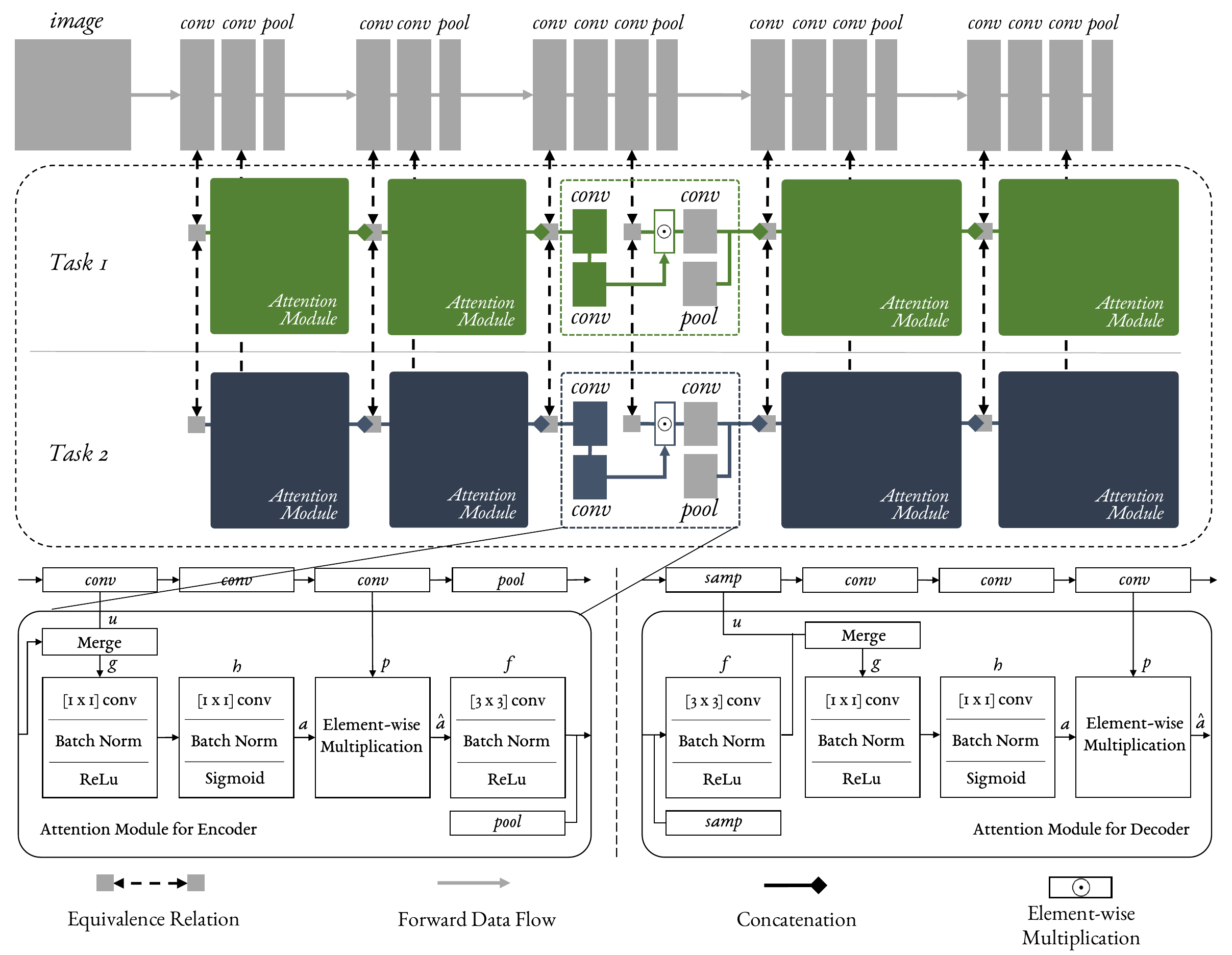}
  \caption{Visualisation of MTAN based on VGG-16, showing the encoder half of SegNet (with the decoder half being symmetrical to the encoder). Task one (green) and task two (blue) have their own set of attention modules, which link with the shared network (grey). The middle attention module has its structure exposed for visualisation, which is further expanded in the bottom section of the figure, showing both the encoder and decoder versions of the module. All attention modules have the same design, although their weights are individually learned.}
  \label{img: mtan}
  \vspace{-0.25cm}
\end{figure*}

MTAN consists of two components: a single shared network, and $K$ task-specific attention networks. The shared network can be designed based on the particular task, whilst each task-specific network consists of a set of attention modules, which link with the shared network. Each attention module applies a soft attention mask to a particular layer of the shared network, to learn task-specific features. As such, the attention masks can be considered as feature selectors from the shared network, which are automatically learned in an end-to-end manner, whilst the shared network learns a compact global feature pool across all tasks.

Figure \ref{img: mtan} shows a detailed visualisation of our network based on VGG-16 \cite{simonyan2014very}, illustrating the encoder half of SegNet. The decoder half of SegNet is then symmetric to VGG-16. As shown, each attention module learns a soft attention mask, which itself is dependent on the features in the shared network at the corresponding layer. Therefore, the features in the shared network, and the soft attention masks, can be learned jointly to maximise the generalisation of the shared features across multiple tasks, whilst simultaneously maximising the task-specific performance due to the attention masks.

\subsection{Task Specific Attention Module}
\label{subsec:attmodule}

The attention module is designed to allow the task-specific network to learn task-related features, by applying a soft attention mask to the features in the shared network, with one attention mask per task per feature channel. We denote the shared features in the $j^{th}$ block of the shared network as $p^{(j)}$, and the learned attention mask in this layer for task $i$ as $a_i^{(j)}$. The task-specific features $\hat{a}^{(j)}_i$ in this layer, are then computed by element-wise multiplication of the attention masks with the shared features:
\begin{align}
  \hat{a}^{(j)}_i &= a_i^{(j)}\odot p^{(j)}~,
\end{align}
where $\odot$ denotes element-wise multiplication.

As shown in Figure \ref{img: mtan}, the first attention module in the encoder takes as input only features in the shared network. But for subsequent attention modules in block $j$, the input is formed by a concatenation of the shared features $u^{(j)}$, and the task-specific features from the previous layer $\hat{a}^{(j-1)}_i$:
\begin{align}
  a_i^{(j)} =
  h_i^{(j)}\left(g_i^{(j)}\left(\left[u^{(j)}; f^{(j)}\left(\hat{a}^{(j-1)}_{i}\right)\right]\right)\right), \, j\geq 2
\end{align}

Here, $f^{(j)},g_i^{(j)}, h_i^{(j)}$ are convolutional layers with batch normalisation, following a non-linear activation. Both $g_i^{(j)}$ and $h_i^{(j)}$ are composed of  $[1\times 1]$ kernels presenting the $i^{th}$ task-specific attention mask in block $j$. $f^{(j)}$ is composed of $[3\times 3]$ kernels representing a shared feature extractor for passing to another attention module, following by a pooling or sampling layer to match the corresponding resolution.

The attention mask, following a sigmoid activation to ensure $a_i^{(j)}\in[0,1]$, is learned in a self-supervised fashion with back-propagation. If $a_i^{(j)}\to 1$ such that the mask becomes an identity map, the attended feature maps are equivalent to global feature maps and the tasks share all the features. Therefore, we expect the performance to be no worse than that of a shared multi-task network, which splits into individual tasks only at the end of the network, and we show results demonstrating this in Section \ref{sec: experiment}.

\subsection{The Model Objective}
\label{subsec: loss}

In general multi-task learning with $K$ tasks, input ${\bf X}$ and task-specific labels ${\bf Y}_i,i=1,2,\cdots,K$, the loss function is defined as,
\begin{align}
\mathcal{L}_{tot}({{\bf X}}, {\bf Y}_{1:K}) = \sum_{i=1}^K\lambda_i\mathcal{L}_i({\bf X},{\bf Y}_i).
\end{align}
This is the linear combination of task-specific losses $\mathcal{L}_i$ with task weightings $\lambda_i$. In our experiments, we study the effect of different weighting schemes on various multi-task learning approaches.

For image-to-image prediction tasks, we consider each mapping from input data ${\bf X}$ to a set of labels ${\bf Y}_i$ as one task with total three tasks for evaluation. In each loss function, $\hat{{\bf Y}}$ represents the network's prediction, and ${\bf Y}$ represents the ground-truth label.

\begin{itemize}[leftmargin=1.2em] 
  \item For semantic segmentation, we apply a pixel-wise cross-entropy loss for each predicted class label from a depth-softmax classifier. 
\begin{align}
  \mathcal{L}_1({\bf X},{\bf Y}_1) = - \frac{1}{pq}\sum_{p,q} {\bf Y}_1(p,q) \log \hat{{\bf Y}}_1(p,q).
\end{align}
 \item For depth estimation, we apply an $L_1$ norm comparing the predicted and ground-truth depth. We use true depth for the NYUv2 indoor scene dataset, and inverse depth in CityScapes outdoor scene dataset as standard, which can more easily represent points at infinite distances, such as the sky:
\begin{align}
  \mathcal{L}_2({\bf X},{\bf Y}_2) =\frac{1}{pq}\sum_{p,q} |{\bf Y}_2(p,q) -\hat{{\bf Y}}_2(p,q)|.
\end{align}
\item For surface normals (only available in NYUv2), we apply an element-wise dot product at each normalised pixel with the ground-truth map:
\begin{align}
  \mathcal{L}_3({\bf X},{\bf Y}_3) = -\frac{1}{pq}\sum_{p,q} {\bf Y}_3(p,q) \cdot \hat{{\bf Y}}_3(p,q).
\end{align}
\end{itemize}

 For image classification tasks, we consider each dataset as one task for which each dataset represents each individual classification task for one domain. We apply standard cross-entropy loss for all classification tasks.

\section{Experiments}
\label{sec: experiment}
In this section, we evaluate our proposed method on two types of tasks: one-to-many predictions for image-to-image regression tasks in Section \ref{subsec: i2ipred} and many-to-many predictions for image classification tasks (Visual Decathlon Challenge) in Section \ref{subsec: visualdecathlon}. 

\subsection{Image-to-Image Prediction (One-to-Many)}
\label{subsec: i2ipred}

In this section, we evaluate MTAN built upon SegNet \cite{badrinarayanan2015segnet} on image-to-image prediction tasks. We first introduce the datasets used for validation in Section \ref{subsec:dataset}, and several baselines for comparison in Section \ref{subsec:network_structure}. In Section \ref{subsec: dwa}, we introduce a novel adaptive weighting method, and in Section \ref{subsec:CityScapes} we show the effectiveness of MTAN with various weighting methods compared with single and multi-task baseline methods. We explore how the performance of our method scales with task complexity in Section \ref{subsec:compare} and we show visualisations of the learned attention masks in Section \ref{subsec:attention}.

\subsubsection{Datasets}
\label{subsec:dataset}

{\bf CityScapes.} The CityScapes dataset \cite{cordts2016CityScapes} consists of high resolution street-view images. We use this dataset for two tasks: semantic segmentation and depth estimation. To speed up training, all training and validation images were resized to $[128\times 256]$. The dataset contains 19 classes for pixel-wise semantic segmentation, together with ground-truth inverse depth labels. We pair the depth estimation task with three levels of semantic segmentation using 2, 7 or 19 classes (excluding the void group in 7 and 19 classes). Labels for the 19 classes and the coarser 7 categories are defined as in the original CityScapes dataset. We then further create a 2-class dataset with only background and foreground objects. The details of these segmentation classes are presented in Table \ref{tab:CityScapes_class}. We perform multi-task learning for 7-class CityScapes dataset in Section \ref{subsec:CityScapes}. We compare the 2/7/19-class results in Section \ref{subsec:compare}, with visualisation of these attention maps in Section \ref{subsec:attention}.

{\bf NYUv2.} The NYUv2 dataset \cite{Silberman:ECCV12} is consisted with RGB-D indoor scene images. We evaluate performances on three learning tasks: 13-class semantic segmentation defined in \cite{couprie2013indoor}, true depth data which is recorded by depth cameras from Microsoft Kinect, and surface normals which are provided in \cite{eigen2015predicting}. To speed up training, all training and validation images were resized to $[288\times 384]$ resolution.

Compared to CityScapes, NYUv2 contains images of indoor scenes, which are much more complex since the viewpoints can vary significantly, changable lighting conditions are present, and the appearance for each object class shifts widely in texture and shape. We evaluate performance on different datasets, together with different numbers of tasks, and further with different class complexities, in order to attain a comprehensive understanding on how our proposed method behaves and scales under a range of scenarios.

\begin{table}[ht!]
  \centering
  \def\arraystretch{1}
  \setlength{\tabcolsep}{0.5em}
  \notsotiny
    \begin{tabular}{cll}
      \toprule
      \multicolumn{1}{c}{\bf 2-class} &\multicolumn{1}{c}{\bf 7-class}  & \multicolumn{1}{c}{\bf 19-class}  \\
      \cmidrule{1-3}
    \multirow{8}[0]{*}[-4pt]{background} & void  & void \\
      \cmidrule{2-3}
          & flat  & road, sidewalk \\
            \cmidrule{2-3}
          & construction & building, wall, fence \\
            \cmidrule{2-3}
          & object & pole, traffic light, traffic sign \\
            \cmidrule{2-3}
          & nature & vegetation, terrain \\
          \cmidrule{2-3}
          & sky   & sky \\
          \cmidrule{1-1}   \cmidrule{2-3}
    \multirow{2}[0]{*}[-2pt]{foreground} & human & person, rider \\
      \cmidrule{2-3}
          & vehicle & carm truck, bus, caravan, trailer, train, motorcycle \\
          \bottomrule
    \end{tabular}%
      \caption{Three  levels of semantic classes for the CityScapes data used in our experiments.}
\label{tab:CityScapes_class}
\vspace{-0.25cm}
\end{table}

\subsubsection{Baselines}
\label{subsec:network_structure}
Most image-to-image multi-task learning architectures are designed based on specific feed-forward neural networks, or implemented on varying network architectures, and thus they are typically not directly comparable based on published results. Our method is general and can be applied to any feed-forward neural network, and so for a fair comparison, we implemented 5 different network architectures (2 single-task + 3 multi-task) based on SegNet \cite{badrinarayanan2015segnet}, which we consider as baselines:

\begin{itemize}[leftmargin=1.2em]
  \setlength\itemsep{0em}
  \item {\bf Single-Task, One Task:} The vanilla SegNet for single task learning.
  \item {\bf Single-Task, STAN:}  A Single-Task Attention Network, where we directly apply our proposed MTAN whilst only performing a single task.
  \item {\bf Multi-Task, Split (Wide, Deep):} The standard multi-task learning, which splits at the last layer for the final prediction for each specific task. We introduce two verions of {\bf Split}: {\bf Wide}, where we adjusted the number of convolutional filters, and {\bf Deep}, where we adjusted the number of convolutional layers, until {\bf Split} had at least as many parameters as MTAN.
  \item {\bf Multi-Task, Dense:} A shared network together with task-specific networks, where each task-specific network receives all features from the shared network, without any attention modules.
  \item {\bf Multi-Task, Cross-Stitch:} The Cross-Stitch Network \cite{misra2016cross}, a previously proposed adaptive multi-task learning approach, which we implemented on SegNet.
\end{itemize}

Note that all the baselines were designed to have at least as many parameters than our proposed MTAN, and were tested to validate that our proposed method's better performance is due to the attention modules, rather than simply due to the increase in network parameters.

\subsubsection{Dynamic Weight Average}
\label{subsec: dwa}
For most multi-task learning networks, training multiple tasks is difficult without finding the correct balance between those tasks, and recent approaches have attempted to address this issue  \cite{chen2017gradnorm,kendall2017multi}. To test our method across a range of weighting schemes, we propose a simple yet effective adaptive weighting method, named Dynamic Weight Average (DWA). Inspired by GradNorm \cite{chen2017gradnorm}, this learns to average task weighting over time by considering the rate of change of loss for each task. But whilst GradNorm requires access to the network's internal gradients, our DWA proposal only requires the numerical task loss, and therefore its implementation is far simpler. 

With DWA, we define the weighting $\lambda_k$ for task $k$ as:
{
\begin{align}
 \lambda_k(t) : = \frac{K\exp(w_k(t-1)/T)}{\sum_i\exp(w_i(t-1)/T)}, w_k(t-1) = \frac{\mathcal{L}_k(t-1)}{\mathcal{L}_k(t-2)},
\end{align}
}%

Here, $w_k(\cdot)$ calculates the relative descending rate in the range $(0,+\infty)$, $t$ is an iteration index, and $T$ represents a temperature which controls the softness of task weighting, similar to \cite{hinton2015distilling}. A large $T$ results in a more even distribution between different tasks. If $T$ is large enough, we have $\lambda_i \approx 1$, and tasks are weighted equally. Finally, the softmax operator, which is multiplied by $K$, ensures that $\sum_i\lambda_i(t)=K$.

In our implementation, the loss value $\mathcal{L}_k(t)$ is calculated as the average loss in each epoch over several iterations. Doing so reduces the uncertainty from stochastic gradient descent and random training data selection. For $t=1,2$, we initialise $w_k(t)=1$, but any non-balanced initialisation based on prior knowledge could also be introduced.

\subsubsection{Results on Image-to-Image Predictions}
\label{subsec:CityScapes}
We now evaluate the performance of our proposed MTAN method in image-to-image multi-task learning, based on the SegNet architecture. Using the 7-class version of the CityScapes dataset and 13-class version of NYUv2 dataset, we compare all the baselines introduced in Section \ref{subsec:network_structure}.

{\bf Training.} For each network architecture, we ran experiments with three types of weighting methods: equal weighting, weight uncertainty \cite{kendall2017multi}, and our proposed DWA (with hyper-parameter temperature $T=2$, found empirically to be optimum across all architectures). We did not include GradNorm \cite{chen2017gradnorm} because it requires a manual choice of subset network weights across all baselines, based on their specific architectures, which distracts from a fair evaluation of the architectures themselves. We trained all the models with ADAM optimiser \cite{kingma2014adam} using a learning rate of $10^{-4}$, with a batch size of 2 for NYUv2 dataset and 8 for CityScapes dataset. During training, we halve the learning rate at 40k iterations, for a total of 80k iterations.

{\bf Results.} Table \ref{tab:mtl_results_city} and \ref{tab:mtl_results_compare} shows experimental results for CityScales and NYUv2 datasets across all architectures, and across all loss function weighting schemes. Results also show the number of network parameters for each architecture. Our MTAN method performs similarly to our baseline Dense in the CityScapes dataset, whilst only having less than half the number of parameters, and outperforms all other baselines. For the more challenging NYUv2 dataset, our method outperforms all baselines across all weighting methods and all learning tasks.

\begin{table}[ht!]
  \centering
  \notsotiny
  \def\arraystretch{0.9}
  \setlength{\tabcolsep}{0.35em}
  \begin{tabularx}{\linewidth}{cll*{4}{c}}
  \toprule
 \multicolumn{1}{c}{\multirow{2.5}[4]{*}{\#P.}} & \multicolumn{1}{c}{\multirow{2.5}[4]{*}{Architecture}} & \multicolumn{1}{c}{\multirow{2.5}[4]{*}{Weighting}} & \multicolumn{2}{c}{Segmentation} & \multicolumn{2}{c}{Depth}  \\
  \cmidrule(lr){4-5} \cmidrule(lr){6-7}
   &\multicolumn{1}{c}{} & \multicolumn{1}{c}{} & \multicolumn{2}{c}{(Higher Better)} & \multicolumn{2}{c}{(Lower Better)} \\
     \multicolumn{1}{c}{} & \multicolumn{1}{c}{} & \multicolumn{1}{c}{} & \multicolumn{1}{c}{mIoU}  & \multicolumn{1}{c}{Pix Acc}  & \multicolumn{1}{c}{Abs Err} & \multicolumn{1}{c}{Rel Err}  \\
\midrule
   2 &   One Task  &n.a. & 51.09 & 90.69 & 0.0158 & 34.17   \\
   3.04 & STAN & n.a.   &  51.90&90.87 & 0.0145 & 27.46 \\
  \midrule
    &   &Equal Weights  & 50.17 & 90.63  & 0.0167  & 44.73  \\
   1.75 & Split, Wide    & Uncert. Weights \cite{kendall2017multi}    & {\bf 51.21} & {\bf 90.72} & {\bf 0.0158} &  44.01   \\
    &    & DWA, $T = 2$ & 50.39 & 90.45 & 0.0164 & {\bf 43.93}     \\
   \cmidrule(lr){1-7}
    &   &Equal Weights  &  {\bf 49.85}  & 88.69 & 0.0180  & 43.86   \\
    2  & Split, Deep    & Uncert. Weights \cite{kendall2017multi}   & 48.12  & 88.68 & {\bf 0.0169} & {\bf 39.73}  \\
    &    & DWA, $T = 2$ & 49.67  & {\bf 88.81}& 0.0182 & 46.63   \\
   \cmidrule(lr){1-7}
    &    & Equal Weights & {\bf 51.91}  & 90.89  & 0.0138 & 27.21  \\
   3.63 & Dense   &Uncert. Weights \cite{kendall2017multi}  &51.89 & {\bf 91.22} & \mybox{\bf 0.0134} & \mybox{\bf 25.36}    \\
    &     &DWA, $T = 2$   & 51.78 & 90.88 & 0.0137 & 26.67     \\
  \cmidrule(lr){1-7}
    &    & Equal Weights &  50.08 & 90.33 & 0.0154 & 34.49    \\
   $\approx$2& Cross-Stitch \cite{misra2016cross}    &Uncert. Weights \cite{kendall2017multi}  & 50.31 & 90.43 & {\bf 0.0152} & {\bf 31.36}    \\
   &     &DWA, $T = 2$   & {\bf 50.33} & {\bf 90.55} & 0.0153 & 33.37 \\
  \cmidrule(lr){1-7}
    &    & Equal Weights   & 53.04 & \mybox{\bf 91.11} &  {\bf 0.0144} & {\bf 33.63}  \\
  1.65 & MTAN (Ours) & Uncert. Weights \cite{kendall2017multi} & \mybox{\bf 53.86} & 91.10  &0.0144  & 35.72  \\
      & &DWA, $T = 2$     & 53.29 & 91.09 & 0.0144  & 34.14 \\
    \bottomrule
    \end{tabularx}%
     \caption{7-class semantic segmentation and depth estimation results on CityScapes validation dataset. \#P shows the number of network parameters, and the best performing combination of multi-task architecture and weighting is highlighted in bold. The top validation scores for each task are annotated with boxes.}
    \label{tab:mtl_results_city}
    \vspace{-0.25cm}
\end{table}

In particular, our method has two key advantages. First, due to the efficiency of having a single shared feature pool with attention masks automatically learning which features to share, our method outperforms other methods without requiring extra parameters (column \#P), and even with significantly fewer parameters in some cases.

\begin{figure}[ht!]
  \centering
  \includegraphics[width=\linewidth]{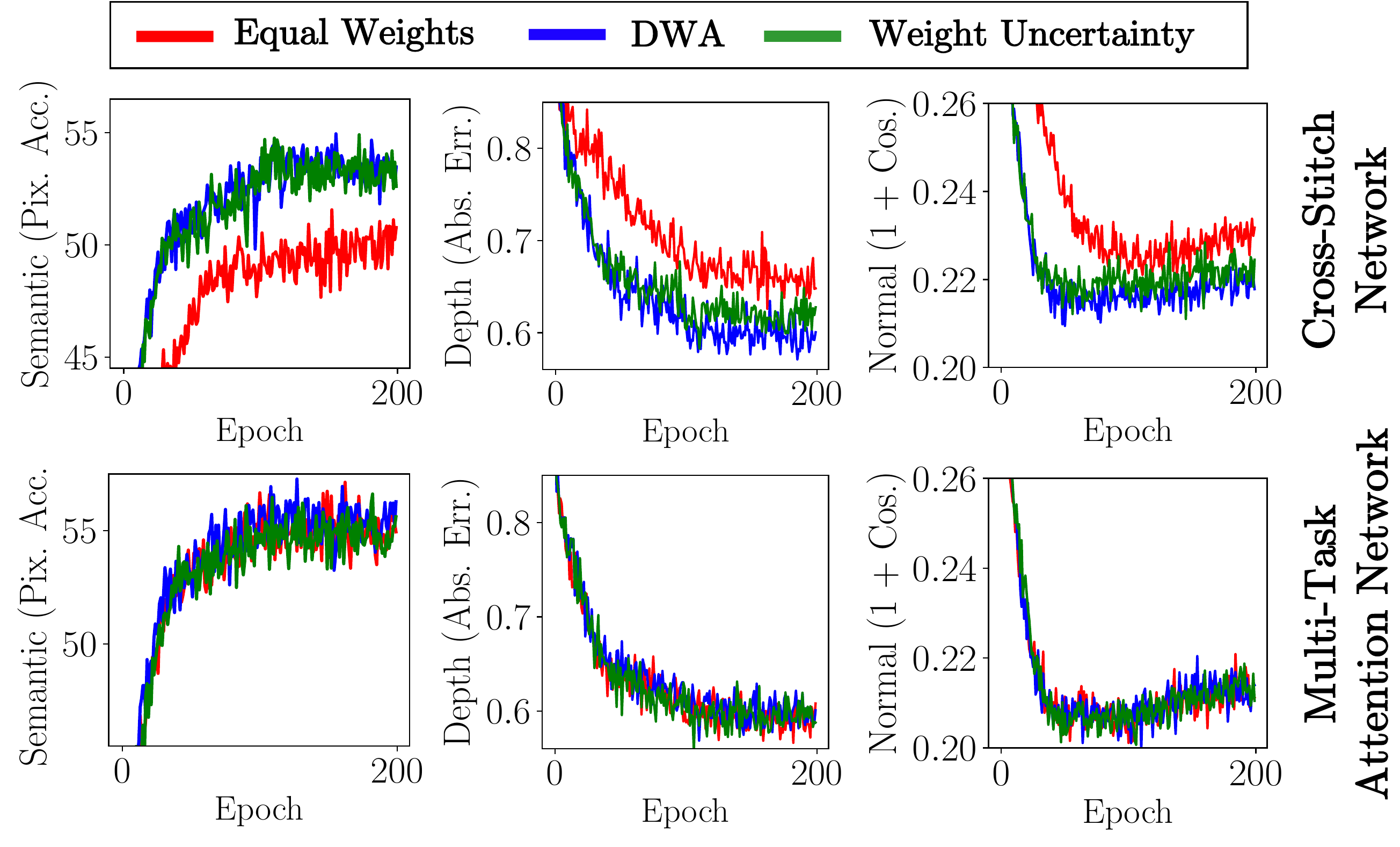}
  \caption{Validation performance curves on the NYUv2 dataset, across all three tasks (semantics, depth, normals, from left to right), showing robustness to loss function weighting schemes on the Cross-Stitch Network \cite{misra2016cross} (top) and our Multi-task Attention Network (bottom).}
  \label{img: loss_compare}
  \vspace{-0.15cm}
\end{figure}

Second, our method maintains high performance across different loss function weighting schemes, and is more robust to the choice of weighting scheme than other methods, avoiding the need for cumbersome tweaking of loss weights. We illustrate the robustness of our method to the weighting schemes with a comparison to the Cross-Stitch Network \cite{misra2016cross}, by plotting learning curves in Figure \ref{img: loss_compare} with respect to the performance of three learning tasks in NYUv2 dataset. We can clearly see that our network follows similar learning trends across various weighting schemes, compared to the Cross-Stitch Network which produces notably different behaviour across the different schemes.

Figure \ref{fig:7class} then shows qualitative results on the CityScapes validation dataset. We can see the advantage of our multi-task learning approach over vanilla single-task learning, where the edges of objects are clearly more pronounced.

\begin{table*}[ht!]
  \centering
  \footnotesize
  \def\arraystretch{0.85}
  \setlength{\tabcolsep}{0.58em}
  \begin{tabularx}{\linewidth}{ccll*{9}{c}}
  \toprule
  \multicolumn{1}{c}{\multirow{3.5}[4]{*}{Type}} & \multicolumn{1}{c}{\multirow{3.5}[4]{*}{\#P.}} & \multicolumn{1}{c}{\multirow{3.5}[4]{*}{Architecture}} & \multicolumn{1}{c}{\multirow{3.5}[4]{*}{Weighting}} & \multicolumn{2}{c}{Segmentation} & \multicolumn{2}{c}{Depth}  & \multicolumn{5}{c}{Surface Normal}\\
  \cmidrule(lr){5-6} \cmidrule(lr){7-8} \cmidrule(lr){9-13}
   &\multicolumn{1}{c}{}  &\multicolumn{1}{c}{} & \multicolumn{1}{c}{} & \multicolumn{2}{c}{\multirow{1.5}[2]{*}{(Higher Better)}} & \multicolumn{2}{c}{\multirow{1.5}[2]{*}{(Lower Better)}}   & \multicolumn{2}{c}{Angle Distance}  & \multicolumn{3}{c}{Within $t^\circ$} \\
    &\multicolumn{1}{c}{}  &\multicolumn{1}{c}{} & \multicolumn{1}{c}{} & \multicolumn{1}{c}{} & \multicolumn{1}{c}{}  & \multicolumn{1}{c}{}  & \multicolumn{1}{c}{} & \multicolumn{2}{c}{(Lower Better)} & \multicolumn{3}{c}{(Higher Better)} \\
     &\multicolumn{1}{c}{} & \multicolumn{1}{c}{} & \multicolumn{1}{c}{} & \multicolumn{1}{c}{mIoU}  & \multicolumn{1}{c}{Pix Acc}  & \multicolumn{1}{c}{Abs Err} & \multicolumn{1}{c}{Rel Err} & \multicolumn{1}{c}{Mean}  & \multicolumn{1}{c}{Median}  & \multicolumn{1}{c}{11.25} & \multicolumn{1}{c}{22.5} & \multicolumn{1}{c}{30} \\
\midrule
  \multirow{2}*{Single Task}  &  3 &   One Task  &n.a.  &  15.10  &51.54  & 0.7508 & 0.3266      &  31.76& 25.51 & 22.12 & 45.33 &  57.13\\
   & 4.56 & STAN & n.a.    &  15.73 & 52.89 &  0.6935 &  0.2891 & 32.09 & 26.32 & 21.49 &44.38  & 56.51\\
  \midrule
  \multirow{15}*{Multi Task}  &  &   &Equal Weights & 15.89  & 51.19   &  0.6494   & 0.2804 & 33.69 & 28.91 & {18.54} & 39.91 & 52.02 \\
  & 1.75 & Split, Wide    & Uncert. Weights \cite{kendall2017multi}  & 15.86  & 51.12  &  {\bf 0.6040} & 0.2570 & {\bf 32.33}  & {\bf 26.62} & {\bf 21.68} & {\bf 43.59} & {\bf 55.36 }\\
  &  &    & DWA, $T = 2$  & {\bf 16.92}  & {\bf 53.72}  &  0.6125  & {\bf 0.2546} & 32.34  &27.10  & 20.69  &  42.73 & 54.74 \\
   \cmidrule(lr){2-13}
    &  &   &Equal Weights & 13.03  & 41.47  &  0.7836 & 0.3326 & 38.28  & 36.55 & 9.50 & 27.11 & 39.63 \\
  & 2 & Split, Deep    & Uncert. Weights \cite{kendall2017multi}  & {\bf 14.53}  & 43.69  & 0.7705  & 0.3340 & {\bf 35.14}  & {\bf 32.13 }& {\bf 14.69} & {\bf 34.52} &  {\bf 46.94 }\\
  &  &    & DWA, $T = 2$  & 13.63  & {\bf 44.41}  & {\bf 0.7581}  & {\bf 0.3227 }& 36.41  & 34.12 & 12.82 & 31.12 & 43.48 \\
   \cmidrule(lr){2-13}
  &  &    &Equal Weights  & 16.06   & 52.73   &  0.6488   & 0.2871  & 33.58 & 28.01 & {20.07} & 41.50 & 53.35 \\
  & 4.95 & Dense   &Uncert. Weights \cite{kendall2017multi}  & {\bf 16.48}  & {\bf 54.40}  &   0.6282  & 0.2761  & {\bf 31.68} & {\bf 25.68}  & {\bf 21.73}  & {\bf 44.58} & {\bf 56.65} \\
  &  &     &DWA, $T = 2$  &  16.15  & 54.35   &  {\bf 0.6059}  &  {\bf 0.2593} &  32.44 & 27.40 & 20.53  & 42.76 & 54.27\\
  \cmidrule(lr){2-13}
  &  &    &Equal Weights   & 14.71  & 50.23   &  0.6481   &  0.2871 & 33.56 & 28.58 & 20.08 & 40.54 & 51.97\\
  &  $\approx$3 & Cross-Stitch \cite{misra2016cross}    &Uncert. Weights \cite{kendall2017multi}  &  15.69 &  52.60  &  0.6277   & 0.2702 & 32.69 &  27.26 & 21.63 &  42.84 &  54.45 \\
  &  &     &DWA, $T = 2$   & {\bf 16.11} &  {\bf  53.19} & {\bf 0.5922}   & {\bf 0.2611} & {\bf 32.34} & {\bf 26.91} & {\bf 21.81} & {\bf 43.14} & {\bf 54.92} \\
  \cmidrule(lr){2-13}
  &  &    & Equal Weights &   \mybox{\bf 17.72} &  55.32    & \mybox{\bf 0.5906}  &  0.2577 & 31.44  & \mybox{\bf 25.37}&  \mybox{\bf 23.17} & 45.65 &57.48\\
  &1.77 & MTAN (Ours) & Uncert. Weights \cite{kendall2017multi}  &  17.67    &  \mybox{\bf 55.61}  &  0.5927    & 0.2592 & \mybox{\bf 31.25} & 25.57 & 22.99 & \mybox{\bf 45.83} & \mybox{\bf 57.67} \\
    &    & &DWA, $T = 2$   &  17.15    &  54.97 &  0.5956  & \mybox{\bf 0.2569} & 31.60 & 25.46 & 22.48 & 44.86 & 57.24 \\
    \bottomrule
    \end{tabularx}%
     \caption{13-class semantic segmentation, depth estimation, and surface normal prediction results on the NYUv2 validation dataset. \#P shows the number of network parameters, and the best performing combination of multi-task architecture and weighting is highlighted in bold. The top validation scores for each task are annotated with boxes.}
    \label{tab:mtl_results_compare}
\end{table*}

\begin{figure*}[ht!]
  \centering
  \footnotesize
  \def\arraystretch{0.85}
  \setlength{\tabcolsep}{0.15em}
    \begin{tabular}{>{\centering\arraybackslash}m{.12\textwidth}m{.17\textwidth}m{.17\textwidth}m{.17\textwidth}m{.17\textwidth}m{.17\textwidth}}
    Input Image &
    \adjincludegraphics[width=\linewidth,trim={0 {.1\width} 0 0},clip]{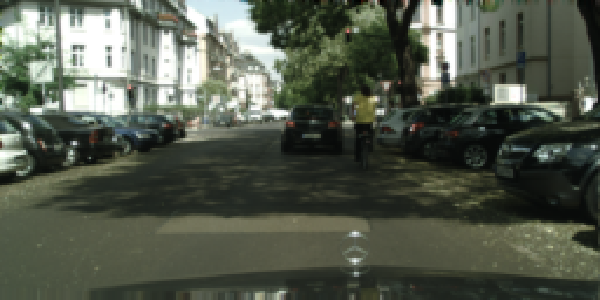}     &
    \adjincludegraphics[width=\linewidth,trim={0 {.1\width} 0 0},clip]{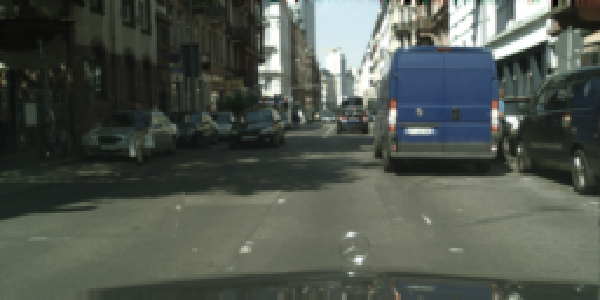}     &
    \adjincludegraphics[width=\linewidth,trim={0 {.1\width} 0 0},clip]{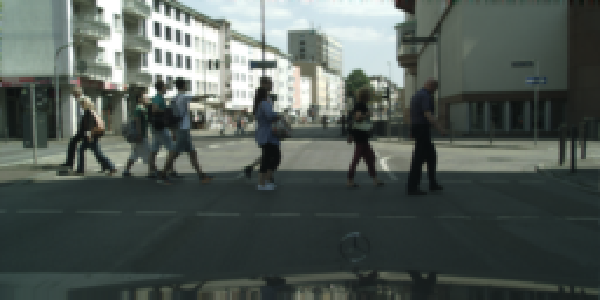} &
    \adjincludegraphics[width=\linewidth,trim={0 {.1\width} 0 0},clip]{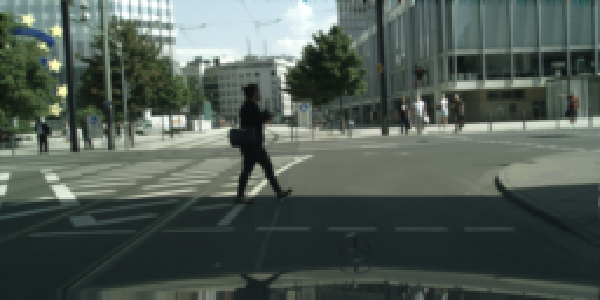}     &
    \adjincludegraphics[width=\linewidth,trim={0 {.1\width} 0 0},clip]{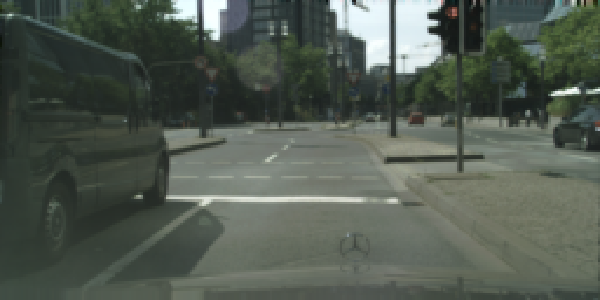}    \\
     \midrule
    Grouth Truth (Semantic) &
    \adjincludegraphics[width=\linewidth,trim={0 {.1\width} 0 0},clip]{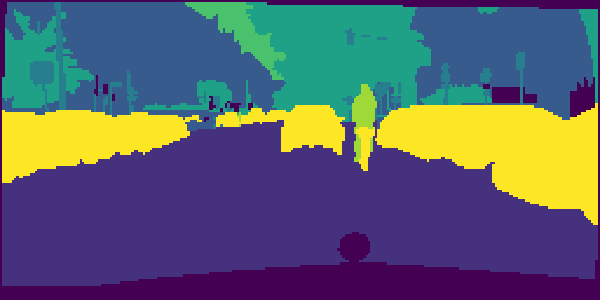}     &
    \adjincludegraphics[width=\linewidth,trim={0 {.1\width} 0 0},clip]{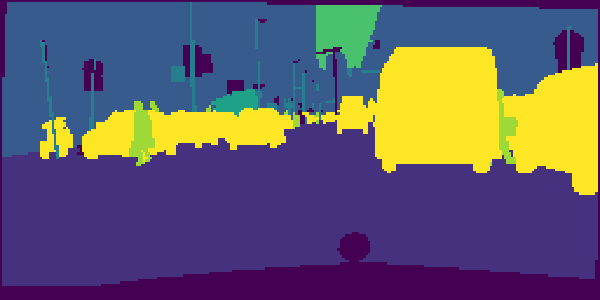}     &
    \adjincludegraphics[width=\linewidth,trim={0 {.1\width} 0 0},clip]{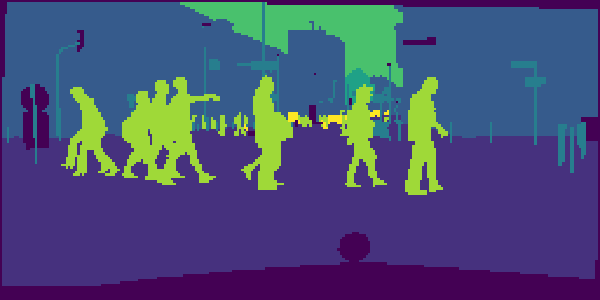} &
    \adjincludegraphics[width=\linewidth,trim={0 {.1\width} 0 0},clip]{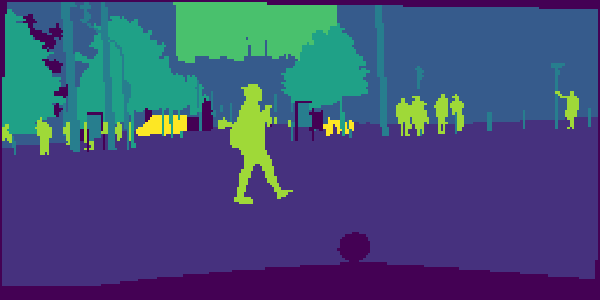}     &
    \adjincludegraphics[width=\linewidth,trim={0 {.1\width} 0 0},clip]{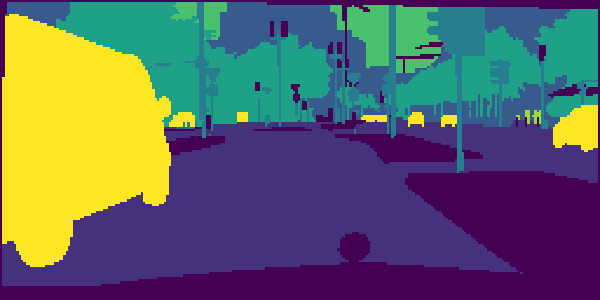}    \\
    Vanilla Single-Task Learning  &
    \adjincludegraphics[width=\linewidth,trim={0 {.1\width} 0 0},clip]{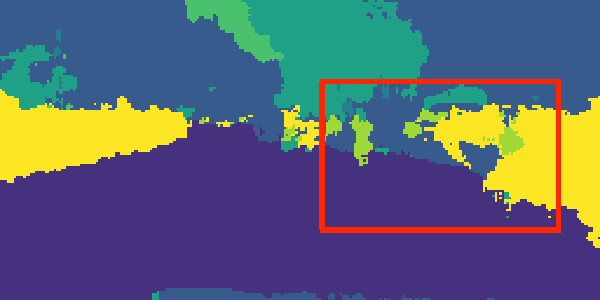}     &
    \adjincludegraphics[width=\linewidth,trim={0 {.1\width} 0 0},clip]{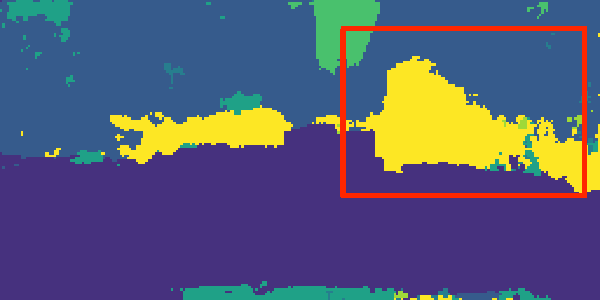}     &
    \adjincludegraphics[width=\linewidth,trim={0 {.1\width} 0 0},clip]{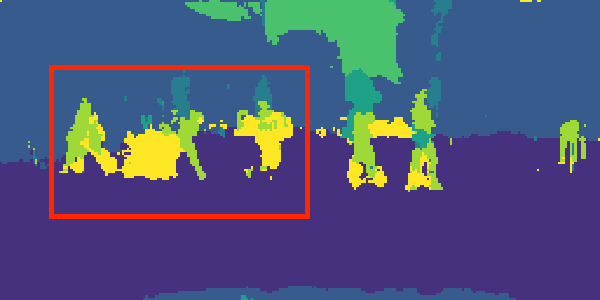} &
    \adjincludegraphics[width=\linewidth,trim={0 {.1\width} 0 0},clip]{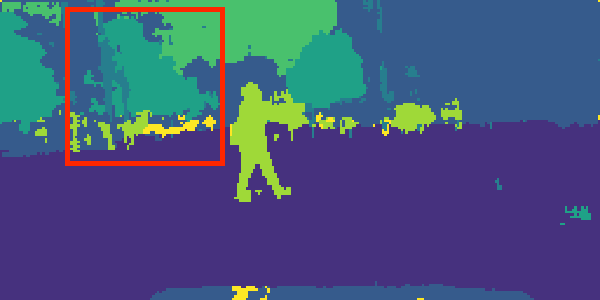}     &
    \adjincludegraphics[width=\linewidth,trim={0 {.1\width} 0 0},clip]{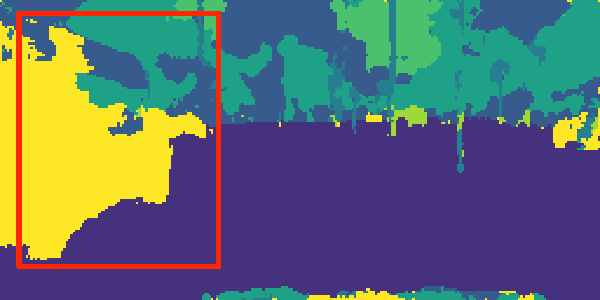}    \\
    Multi-Task Attention Network &
    \adjincludegraphics[width=\linewidth,trim={0 {.1\width} 0 0},clip]{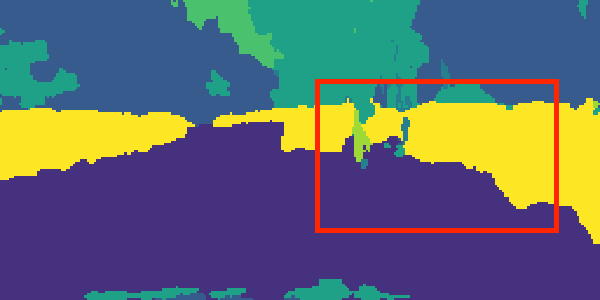}     &
    \adjincludegraphics[width=\linewidth,trim={0 {.1\width} 0 0},clip]{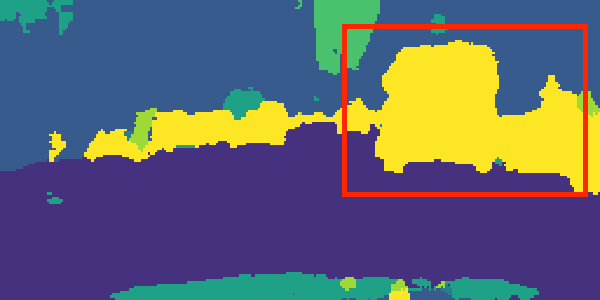}     &
    \adjincludegraphics[width=\linewidth,trim={0 {.1\width} 0 0},clip]{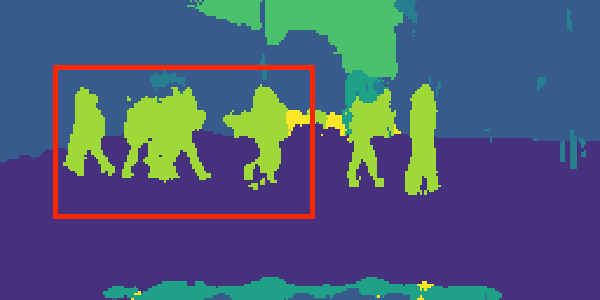} &
    \adjincludegraphics[width=\linewidth,trim={0 {.1\width} 0 0},clip]{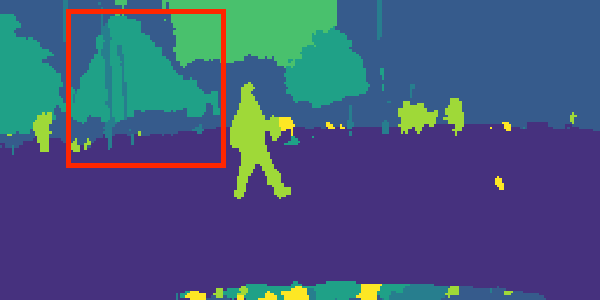}     &
    \adjincludegraphics[width=\linewidth,trim={0 {.1\width} 0 0},clip]{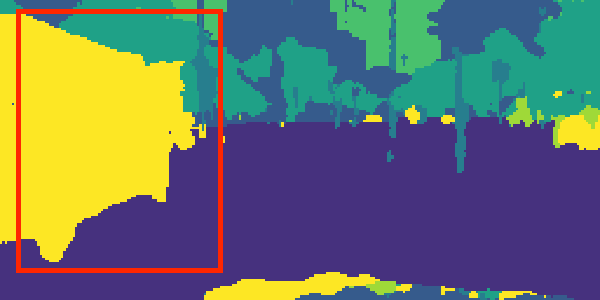}   \\
    \midrule
    Grouth Truth (Depth) &
    \adjincludegraphics[width=\linewidth,trim={0 {.1\width} 0 0},clip]{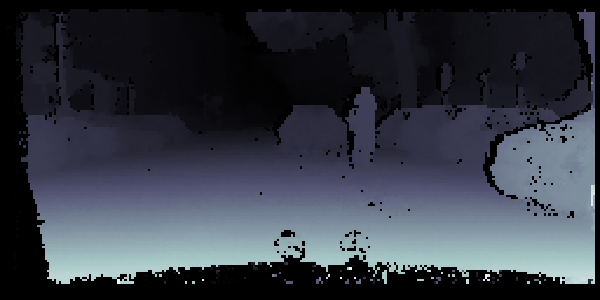}     &
    \adjincludegraphics[width=\linewidth,trim={0 {.1\width} 0 0},clip]{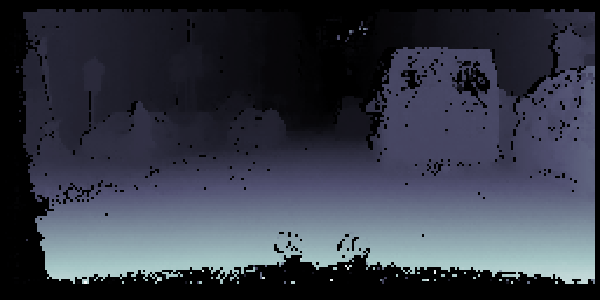}     &
    \adjincludegraphics[width=\linewidth,trim={0 {.1\width} 0 0},clip]{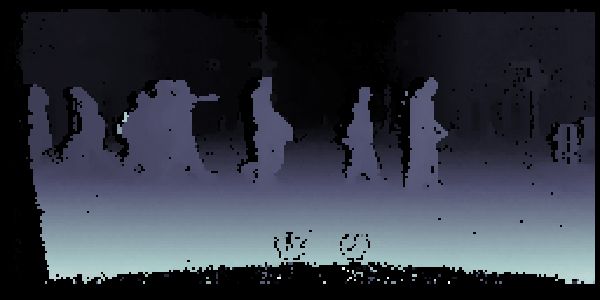} &
    \adjincludegraphics[width=\linewidth,trim={0 {.1\width} 0 0},clip]{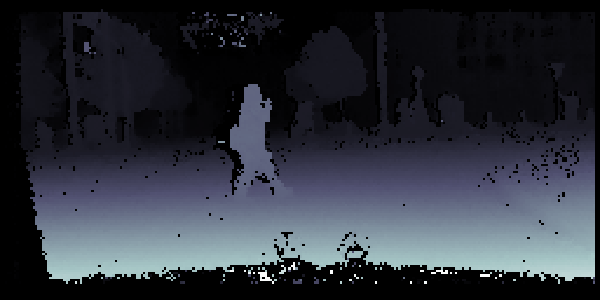} &
    \adjincludegraphics[width=\linewidth,trim={0 {.1\width} 0 0},clip]{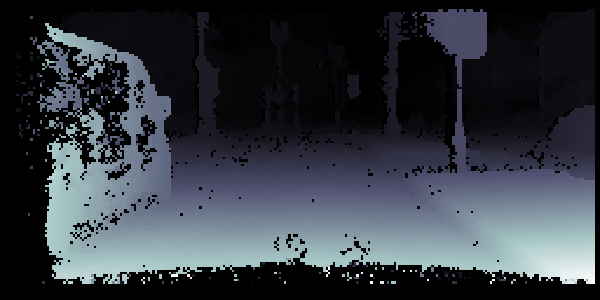}    \\
    Vanilla Single-Task Learning  &
    \adjincludegraphics[width=\linewidth,trim={0 {.1\width} 0 0},clip]{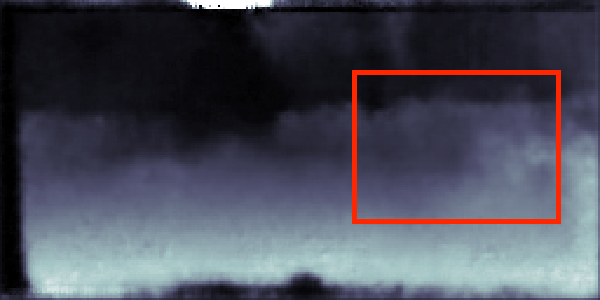}     &
    \adjincludegraphics[width=\linewidth,trim={0 {.1\width} 0 0},clip]{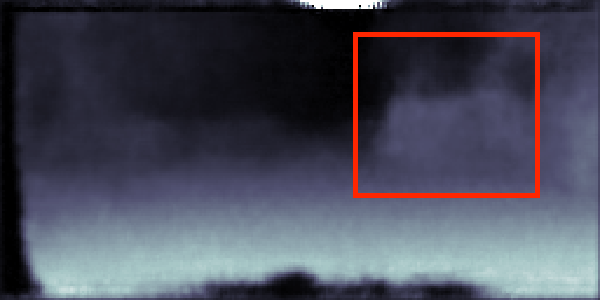}     &
    \adjincludegraphics[width=\linewidth,trim={0 {.1\width} 0 0},clip]{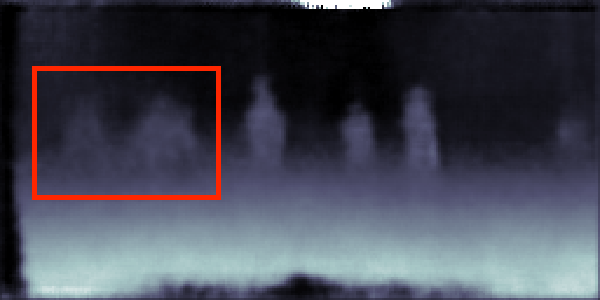} &
    \adjincludegraphics[width=\linewidth,trim={0 {.1\width} 0 0},clip]{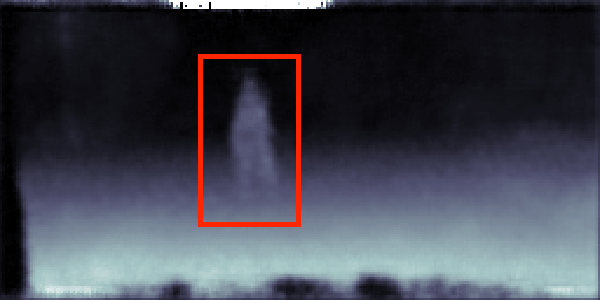}     &
    \adjincludegraphics[width=\linewidth,trim={0 {.1\width} 0 0},clip]{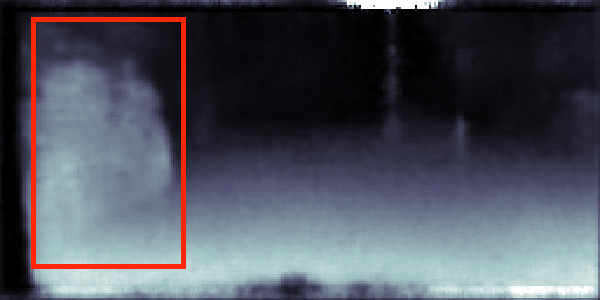}    \\
    Multi-Task Attention Network&
    \adjincludegraphics[width=\linewidth,trim={0 {.1\width} 0 0},clip]{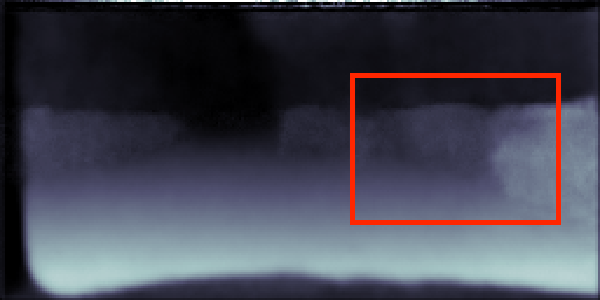}     &
    \adjincludegraphics[width=\linewidth,trim={0 {.1\width} 0 0},clip]{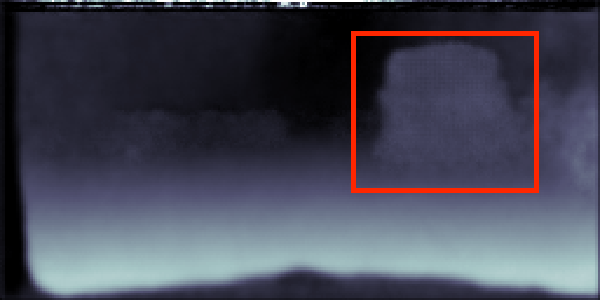}     &
    \adjincludegraphics[width=\linewidth,trim={0 {.1\width} 0 0},clip]{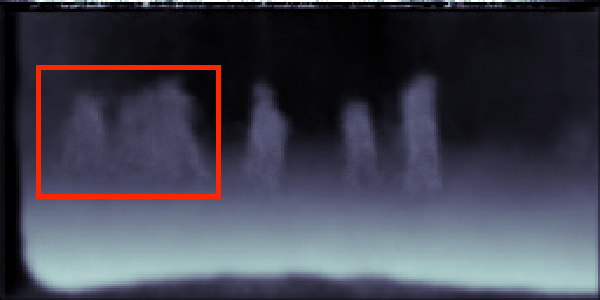} &
    \adjincludegraphics[width=\linewidth,trim={0 {.1\width} 0 0},clip]{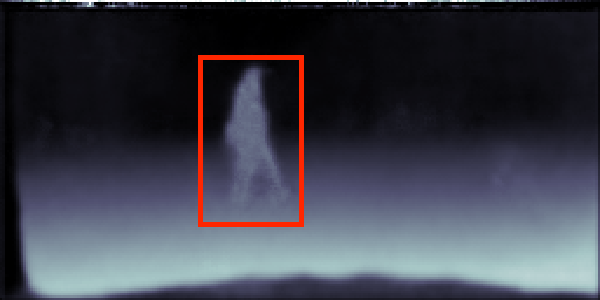}     &
    \adjincludegraphics[width=\linewidth,trim={0 {.1\width} 0 0},clip]{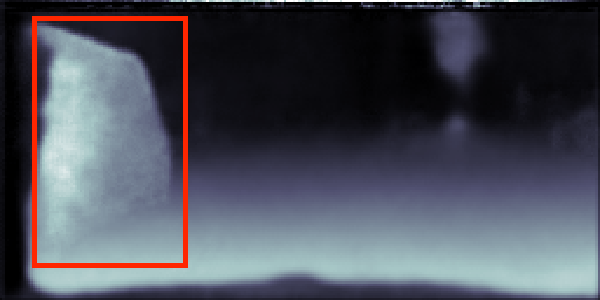}    \\
    \end{tabular}%
\caption{CityScapes validation results on 7-class semantic labelling and depth estimation, trained with equal weighting. The original images are cropped to avoid invalid points for better visualisation. The red boxes are regions of interest, showing the effectiveness of the results provided from our method and single task method.}
\label{fig:7class}
\vspace{-0.25cm}
\end{figure*}

\subsubsection{Effect of Task Complexity}
\label{subsec:compare}

For further introspection into the benefits of multi-task learning, we evaluated our implementations on CityScapes across different numbers of semantic classes, with the depth labels the same across all experiments. We trained the networks with the same settings as in Section \ref{subsec:CityScapes}, with an additional multi-task baseline {\bf Split} (the standard version), which we found to perform better than the other modified versions. All networks are trained with equal weighting.

\begin{table*}[ht!]
  \begin{minipage}{.28\textwidth}
  \centering
  \includegraphics[width=4.5cm,clip]{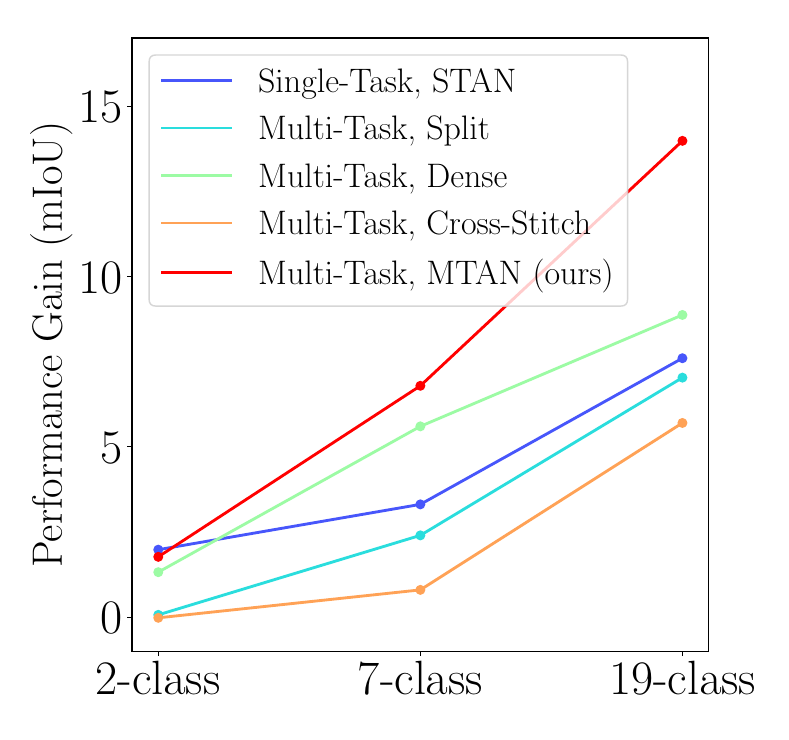}
  \def\baselinestretch{0.85}
  \label{fig:konvpar}
  \end{minipage}
  \begin{minipage}{.7\textwidth}
  \centering
  \def\arraystretch{1.1}
  \setlength{\tabcolsep}{0.18em}
  \footnotesize
  \begin{tabular}{l|c|cccccccccc|c|c}
    \toprule
      {\bf Method} & {\bf \#P.} & {\bf ImNet.} & {\bf Airc.} & {\bf C100} & {\bf DPed} & {\bf DTD} & {\bf GTSR} & {\bf Flwr} & {\bf Oglt} & {\bf SVHN} & {\bf UCF} & {\bf Mean} & {\bf Score}\\
      \midrule
      Scratch~\cite{rebuffi2017learning} & 10 & 59.87 & 57.10 & 75.73 & 91.20 & 37.77 & 96.55 & 56.3 & 88.74 & 96.63 & 43.27 & 70.32 & 1625 \\
      Finetune~\cite{rebuffi2017learning} & 10 & 59.87 & 60.34 & 82.12 & 92.82 & 55.53 & 97.53 & 81.41 & 87.69 & 96.55 & 51.20 & 76.51 & 2500  \\
      \midrule
      Feature~\cite{rebuffi2017learning} & 1 & 59.67 & 23.31 & 63.11 & 80.33 & 45.37 & 68.16 & 73.69 & 58.79 & 43.54 & 26.8 & 54.28 & 544   \\
      Res. Adapt.\cite{rebuffi2017learning} & 2 & 59.67 & 56.68 & 81.20 & 93.88 & 50.85 & 97.05 & 66.24 & 89.62 & 96.13 & 47.45 & 73.88 & 2118 \\
      DAN~\cite{rosenfeld2017incremental} & 2.17 & 57.74 & 64.12 & 80.07 & 91.30 & 56.54 & 98.46 & 86.05 & 89.67 & 96.77 & 49.38 & 77.01 & 2851 \\
       Piggyback~\cite{mallya2018piggyback} & 1.28 & 57.69 & 65.29 & 79.87 & 96.99 & 57.45 & 97.27 & 79.09 & 87.63 & 97.24 & 47.48 & 76.60 & 2838 \\
      Parallel SVD~\cite{rebuffi2018efficient} & 1.5 & 60.32 & 66.04 & 81.86 & 94.23 & 57.82 & 99.24 & 85.74 & 89.25 & 96.62 & 52.50 & 78.36 & 3398 \\
      MTAN (Ours)  & 1.74 & 63.90   &  61.81  &  81.59   &  91.63  & 56.44  & 98.80  & 81.04  & 89.83   & 96.88  & 50.63  & 77.25 & 2941 \\
      \bottomrule
    \end{tabular}
  \end{minipage}%
  \caption{Left: CityScapes performance gain in percentage for all implementations compared with the vanilla single-task method. Right: Top-1 classification accuracy on the Visual Decathlon Challenge online test set. \#P is the number of parameters as a factor of a single-task implementation. The upper part of table presents results from single task learning baselines; lower part of table presents results from multi-task learning baselines.}
  \label{img: performance}
  \vspace{-0.25cm}
  \end{table*}

Table \ref{img: performance} (left) shows the validation performance improvement across all multi-task implementations and the single-task STAN implementation, plotted relative to the performance of the vanilla single-task learning on the CityScapes dataset. Interestingly, for only a 2-class setup, the single-task attention network (STAN) performs better than all multi-task methods since it is able to fully utilise network parameters in a simple manner for the simple task. However, for greater task complexity, the multi-task methods encourage the sharing of features for a more efficient use of available network parameters, which then leads to better results. We also observe that, whilst the relative performance gain increases for all implementations as the task complexity increases, our MTAN method increases at a greater rate.

\subsubsection{Attention Masks as Feature Selectors}
\label{subsec:attention}

\begin{figure}[ht!]
  \def\arraystretch{0.8}
  \setlength{\tabcolsep}{0.15em}
  \centering
  \footnotesize
    \begin{tabular}{>{\centering\arraybackslash}m{.32\linewidth}m{.32\linewidth}m{.32\linewidth}}
    \multicolumn{1}{c}{Input Image}  & \multicolumn{1}{c}{Semantic Mask} & \multicolumn{1}{c}{Semantic Features} \\
    \adjincludegraphics[width=\linewidth,trim={0 {.1\width} 0 0},clip]{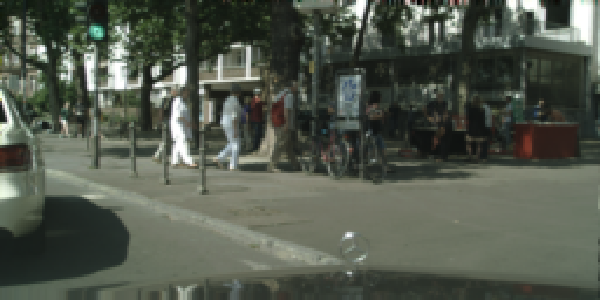}    &
    \adjincludegraphics[width=\linewidth,trim={0 {.1\width} 0 0},clip]{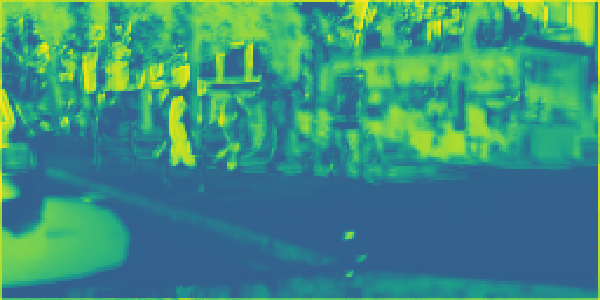}     &
    \adjincludegraphics[width=\linewidth,trim={0 {.1\width} 0 0},clip]{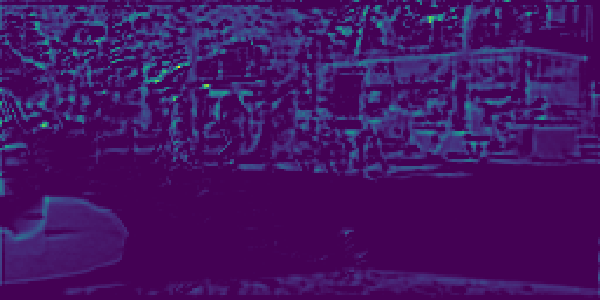}   \\
    \multicolumn{1}{c}{Shared Features}  & \multicolumn{1}{c}{Depth Mask} & \multicolumn{1}{c}{Depth Features} \\
    \adjincludegraphics[width=\linewidth,trim={0 {.1\width} 0 0},clip]{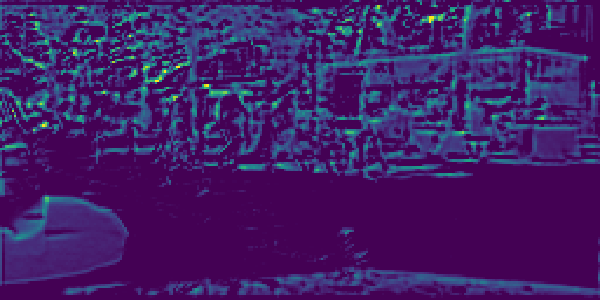}     &
    \adjincludegraphics[width=\linewidth,trim={0 {.1\width} 0 0},clip]{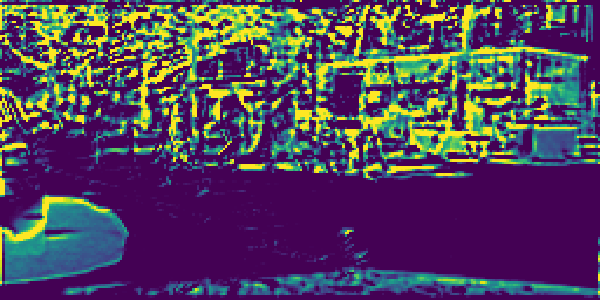}     &
    \adjincludegraphics[width=\linewidth,trim={0 {.1\width} 0 0},clip]{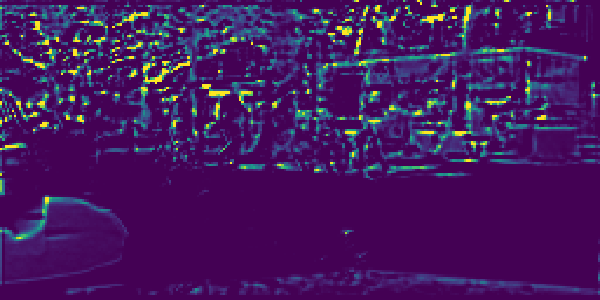} \\
    \midrule
    \multicolumn{1}{c}{Input Image}  & \multicolumn{1}{c}{Semantic Mask} & \multicolumn{1}{c}{Semantic Features} \\
    \adjincludegraphics[width=\linewidth,trim={0 {.1\width} 0 0},clip]{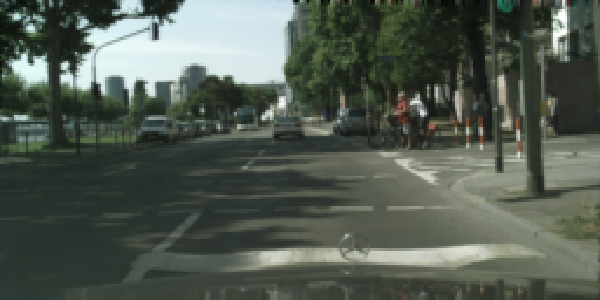}    &
    \adjincludegraphics[width=\linewidth,trim={0 {.1\width} 0 0},clip]{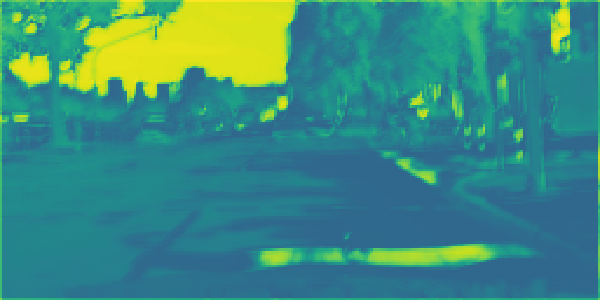}     &
    \adjincludegraphics[width=\linewidth,trim={0 {.1\width} 0 0},clip]{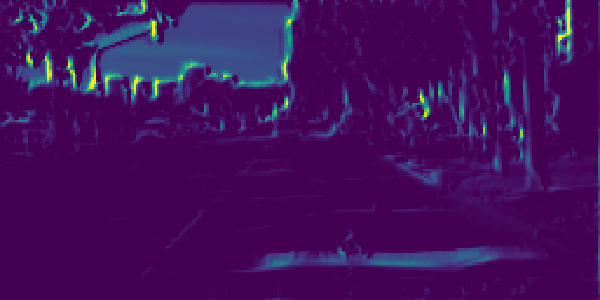}   \\
     \multicolumn{1}{c}{Shared Features}  & \multicolumn{1}{c}{Depth Mask} & \multicolumn{1}{c}{Depth Features} \\
    \adjincludegraphics[width=\linewidth,trim={0 {.1\width} 0 0},clip]{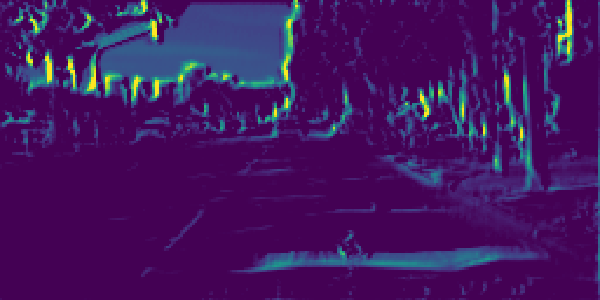}     &
    \adjincludegraphics[width=\linewidth,trim={0 {.1\width} 0 0},clip]{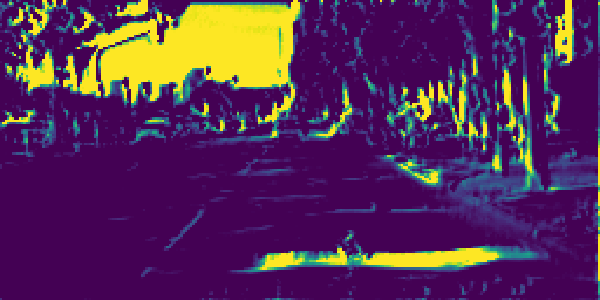}     &
    \adjincludegraphics[width=\linewidth,trim={0 {.1\width} 0 0},clip]{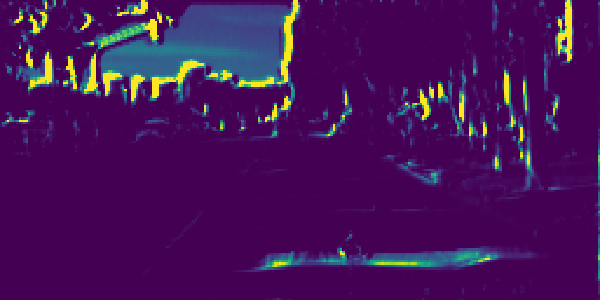} \\
    \end{tabular}%
\caption{Visualisation of the first layer of 7-class semantic and depth  attention features of our proposed network. The colours for each image are rescaled to fit the data.}
\label{fig:attention}
\vspace{-0.25cm}
\end{figure}

To understand the role of the proposed attention modules, in Figure \ref{fig:attention} we visualise the first layer attention masks learned with our network based on CityScapes dataset. We can see a clear difference in attention masks between the two tasks, with each mask working as a feature selector to mask out uninformative parts of the shared features, and focus on parts which are useful for each task. Notably, the depth masks have a much higher contrast than the semantic masks, suggesting that whilst all shared features are generally useful for the semantic task, the depth task benefits more from extraction of task-specific features.

\subsection{Visual Decathlon Challenge (Many-to-Many)}
\label{subsec: visualdecathlon}

Finally, we evaluate our approach on the recently introduced Visual Decathlon Challenge, consisting of 10 individual image classification tasks (many-to-many predictions). Evaluation on this challenge reports per-task accuracies, and assigns a cumulative score with a maximum value of 10,000 (1,000 per task) based on these accuracies. The complete details about the challenge settings, evaluation, and datasets used, can be found at \url{http://www.robots.ox.ac.uk/~vgg/decathlon/. }

Table \ref{img: performance} (right) shows results for the online test set of the challenge. As consistent with the prior works, we apply MTAN built on Wide Residual Network \cite{BMVCwrn} with a depth of 28, widening factor of 4, and a stride of 2 in the first convolutional layer of each block. We train our model using a batch size of 100, learning rate of 0.1 with SGD, and weight decay of $5\cdot 10^{-5}$ for all 10 classification tasks. We halve the learning rate every 50 epochs for a total of 300 epochs. Then, we fine-tune 9 classification tasks (all except ImageNet) with a learning rate 0.01 until convergence. The results show that our approach surpasses most of the baselines and is competitive with the current state-of-the-art, without the need for complicated regularisation strategies such as applying DropOut \cite{srivastava2014dropout}, regrouping datasets by size, or adaptive weight decay for each dataset, as required.

\section{Conclusions}

In this work, we have presented a new method for multi-task learning, the Multi-Task Attention Network (MTAN). The network architecture consists of a global feature pool, together with task-specific attention modules for each task, which allows for automatic learning of both task-shared and task-specific features in an end-to-end manner. Experiments on the NYUv2 and CityScapes datasets with multiple dense-prediction tasks, and on the Visual Decathlon Challenge with multiple image classification tasks, show that our method outperforms or is competitive with other methods, whilst also showing robustness to the particular task weighting schemes used in the loss function. Due to our method's ability to share weights through attention masks, our method achieves this state-of-the-art performance whilst also being highly parameter efficient.

{\small
\bibliographystyle{ieee_fullname}
\bibliography{egbib}
}

\end{document}